\documentclass[runningheads]{llncs}
\usepackage{graphicx}
\usepackage[ruled]{algorithm2e}
\usepackage{tikz}
\usepackage{comment}
\usepackage{amsmath,amssymb} 
\usepackage{color}
\usepackage{multirow}
\usepackage{booktabs}
\usepackage{hyperref}

\usepackage[accsupp]{axessibility}  

\begin{document}
\pagestyle{headings}
\mainmatter
\def\ECCVSubNumber{1616}  

\title{Watermark Vaccine: Adversarial Attacks to Prevent Watermark Removal} 

\titlerunning{Adversarial Attacks to Prevent Watermark Removal}
%
\author{Xinwei Liu$^{1,2}$, 
Jian Liu $^{3}$, 
Yang Bai$^{4}$ ,
Jindong Gu$^{5}$,
Tao Chen$^{3}$,\\
Xiaojun Jia$^{1,2}$ \thanks{Corresponding Author}, 
Xiaochun Cao$^{1,6}$
}
\authorrunning{Xinwei Liu et al.}
%
\institute{
$^{1}$ SKLOIS, Institute of Information Engineering, CAS, Beijing, China\\
$^{2}$ School of Cyber Security, University of Chinese Academy of Sciences, Beijing, China\\
$^{3}$ Ant Group, Beijing, China \\
$^{4}$ Tencent Security Zhuque Lab, Beijing, China\\
$^{5}$ University of Munich, Munich, Germany\\
$^{6}$ School of Cyber Science and Technology, Shenzhen Campus, Sun Yat-sen University, Shenzhen 518107, China\\
\email{\{liuxinwei, jiaxiaojun\}@iie.ac.cn \; \{rex.lj,boshan.ct\}@antgroup.com \; mavisbai@tencent.com \; jindong.gu@outlook.com \; caoxiaochun@mail.sysu.edu.cn}
}

\maketitle
\begin{abstract}

As a common security tool, visible watermarking has been widely applied to protect copyrights of digital images. However, recent works have shown that visible watermarks can be removed by DNNs without damaging their host images. Such watermark-removal techniques pose a great threat to the ownership of images. Inspired by the vulnerability of DNNs on adversarial perturbations, we propose a novel defence mechanism by adversarial machine learning for good. 
From the perspective of the adversary, blind watermark-removal networks can be posed as our target models; then we actually optimize an imperceptible adversarial perturbation on the host images to proactively attack against watermark-removal networks, dubbed \textbf{\textit{Watermark Vaccine}}.
Specifically, two types of vaccines are proposed. Disrupting Watermark Vaccine (DWV) induces to ruin the host image along with watermark after passing through watermark-removal networks. In contrast, Inerasable Watermark Vaccine (IWV) works in another fashion of trying to keep the watermark not removed and still noticeable. Extensive experiments demonstrate the effectiveness of our DWV/IWV in preventing watermark removal, especially on various watermark removal networks. The Code is released in \href{github}{https://github.com/thinwayliu/Watermark-Vaccine}.

\end{abstract}

\section{Introduction}

\begin{figure}[t]
\centering
\includegraphics[width=\linewidth]{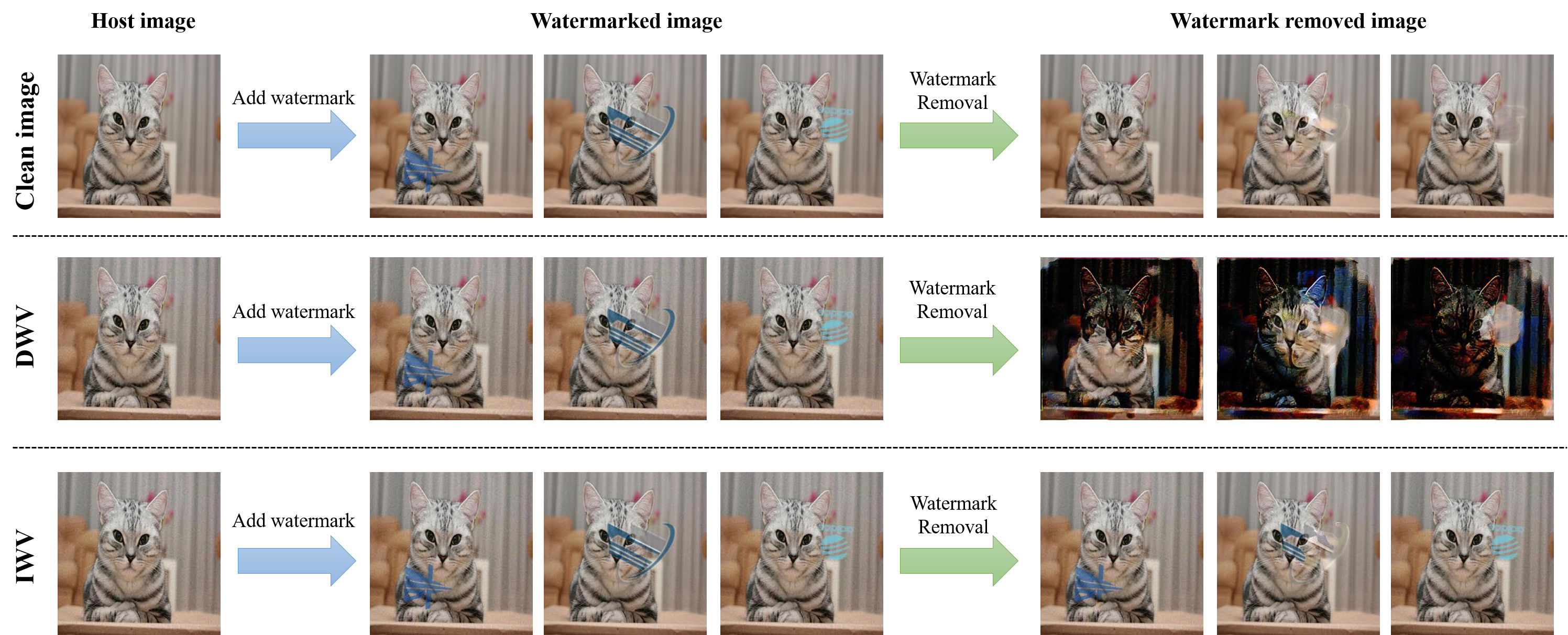}
\caption{The protective effects of our watermark vaccines on different watermark patterns or parameters. The current blind watermark-removal technique, such as WDNet\cite{liu2021wdnet}, can effectively remove the watermarks (\textbf{top}). When the host images are equipped with Disrupting Watermark Vaccine (DWV), the watermark-removed images will be ruined (\textbf{middle}). However, when the host images are equipped with Inerasable Watermark Vaccine (IWV), the results can not be purified successfully as the host images (\textbf{bottom}).}
\label{fig1}
\end{figure}
With the rapid development of digital media and the increasing dependence of deep neural networks (DNNs) on enormous training data, copyright protection attracts great attention especially for image data~\cite{samuel2004digital}. Visible watermarking thus becomes an essential technique~\cite{Braudaway97protecting}. It prevents illicit users from obtaining some critical information and using copyrighted high-quality images. As a result, it can reduce illegal theft and play a role in publicity and warning. Mintzer \emph{et al.}~\cite{mintzer1997effective} posed two characteristics for visible watermark, that is, it can be recognized by human eyes but will not significantly obscure the details.

\begin{figure*}[htb]
\centering
\includegraphics[width=1\textwidth]{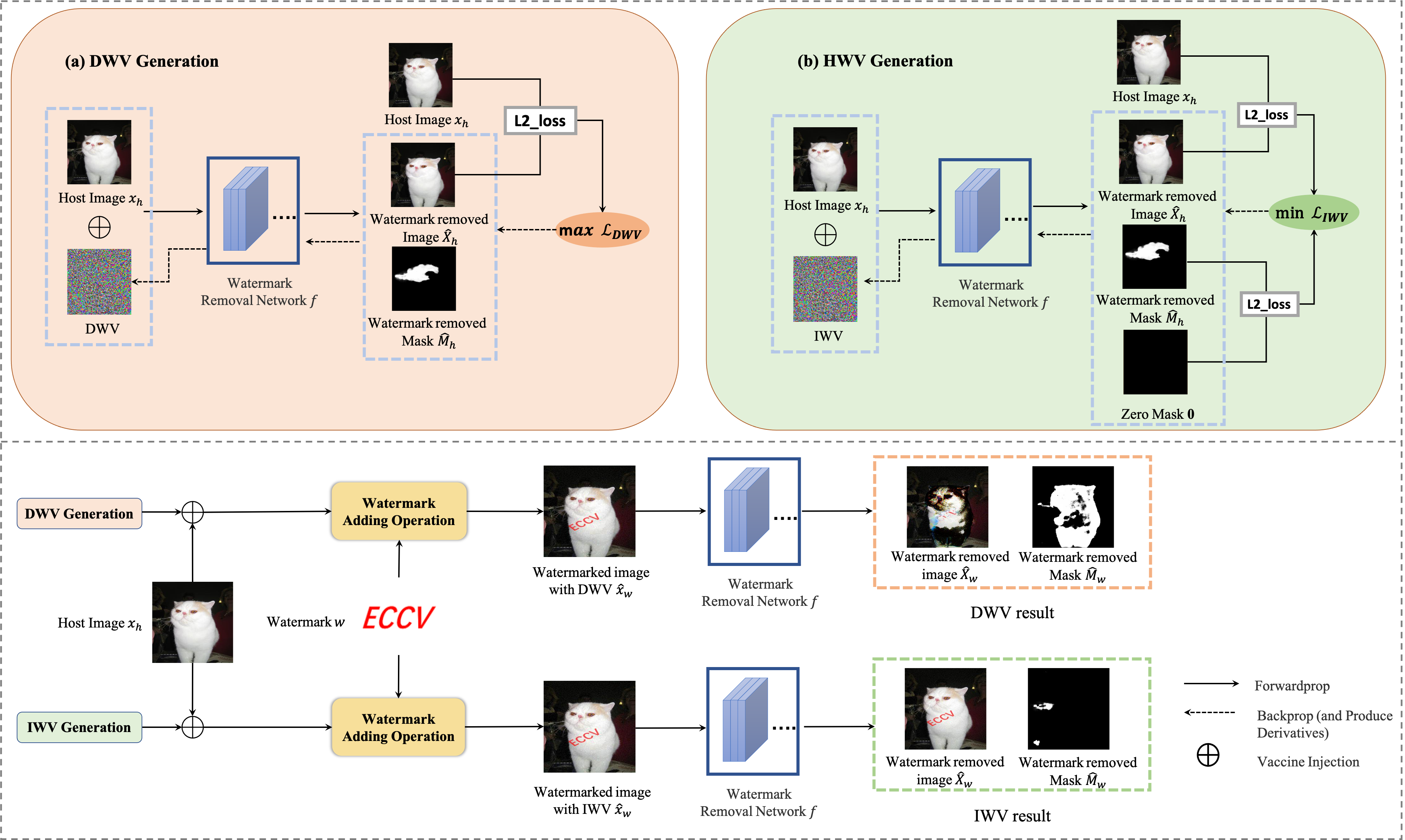}
\caption{The overview figure of the generation (on the first row) and application (on the second row) of our proposed watermark vaccine. We propose to maximize $\mathcal{L_{DWV}}$ to generate DWV and minimize $\mathcal{L_{IWV}}$ to generate IWV, as shown in (a) and (b), respectively. Then we first apply DWV/IWV on the host images to generate 'protected host images'. When watermarks are added on those protected host images, they are hard to be normally removed, tending to show either ruined or watermark-preserving results.}
\vspace{-4mm}
\label{framework}
\end{figure*}

However, visible watermark is in face of security issues as it can be effectively removed by some watermark-removal techniques~\cite{cox2007digital,bertalmio2000image,qin2018visible,santoyo2017automatic,xu2017automatic}. Among these techniques, some require the location area of the watermark. Huang \emph{et al.}~\cite{huang2004attacking} propose to remove the watermark using image inpainting. Park \emph{et al.}~\cite{park2012identigram} propose to formulate it into a feature matching problem. With the rapid development of deep learning, the community has proposed some blind visible watermark-removal DNNs, which can reconstruct watermarked images end-to-end without any information about watermarks~\cite{cao2019generative,cheng2018large,cun2021split,hertz2019blind,li2019towards,liang2021visible,liu2021wdnet}. These works usually adopt two-stage strategies. In the first stage, the networks aim to predict the watermark region. After that, in the second stage, they work on recovering the background of such a watermark region. Without strong assumptions, the blind watermark-removal network has become a mainstream method. In Figure~\ref{fig1} (a), we take WDNet~\cite{liu2021wdnet} as an example to show its performance in both identifying and removing the watermark.

Due to these advanced watermark-removal technologies, traditional watermarking methods can no longer effectively protect the copyrights of picture owners. Recently, Khachaturov \emph{et al.}~\cite{khachaturov2021markpainting} proposed to fool the inpainting-based removal networks to protect the watermark. However, this type of networks are demanding and not widely used. So we focus on preventing the blind watermark removal networks in this paper. Inspired by recent studies on adversary~\cite{akhtar2018threat,carlini2017towards,szegedy2013intriguing,goodfellow2014explaining,yuan2019adversarial,jia2020adv}, which show that imperceptible adversarial perturbations can cause some incorrectly outputs for DNNs, `adversarial for good' is thus a new protection method.
To note that, generating a simple adversarial perturbation on the watermarked image could not work directly. Because in real-world scenarios, a watermark is always automatically generated at the last step by the system or website, and the protected image is required to finish uploading before it. More importantly, the watermark is not permanent for one host image. The watermark could be changed with some specific circumstances (such as Enterprise renaming, logo changes, year updates, etc.). So it can be costly to regenerate an adversarial perturbation for the same host image yet with a different watermark. Thus, a universal perturbation can be useful and efficient for watermark protection.

In this paper, we propose a watermark-agnostic perturbation against blind watermark-removal network, dubbed \textit{\textbf{Watermark Vaccine}}, which is injected on host images before adding watermark just like vaccination in reality. Our method is equivalent to a `double insurance' for copyright protection: the visible watermark serves as a warning and annotating function, telling people not to infringe. The watermark vaccine ensures that the visible watermark won't be removed by blind watermark-removal networks, which can effectively reduce illegal dissemination and other infringements. Then the vaccinated image can protect any watermark from being removed. Specifically, we propose two types of watermark vaccines according to their attack effects: Disrupting Watermark Vaccine (DWV) and Inerasable Watermark Vaccine (IWV). DWV aims to disrupt the watermark-removed image while IWV attempts to still keep the watermark through blind watermark-removal networks. The framework of our proposed watermark vaccine’s generation and is shown in Fig.~\ref{framework}. In Fig.~\ref{fig1} (b) and (c), we can see that after injected with DWV or IWV, the watermark-removed images will either be ruined or the watermarks are not completely removed. In addition, the masks are disrupted for DWV or are induced elsewhere for IWV. Both results demonstrate the effectiveness of our proposed watermark vaccine in successfully preventing watermark removal. 

Our key contributions are summarized as follows:
\begin{itemize}
\item We are the first to propose the watermark-agnostic perturbations for blind watermark-removal networks, dubbed \textit{Watermark Vaccine}, to prevent the watermark removal from host images.
\item We present two types of effective and powerful watermark vaccines (DWV and IWV), which aim to either disrupt the watermark-removed images or keep the watermarks uncleared respectively.
\item We evaluate the effectiveness and universality of two vaccines. The results demonstrate that they generalize well on different watermark patterns, sizes, locations as well as transparencies. In addition, our watermark vaccine can also resist some common image processing operations.
\end{itemize}

\section{Related Work}
\noindent \textbf{Visible Watermark Removal}. Visible watermark-removal techniques are proposed to evaluate and improve the robustness of visible watermarks at the beginning. In the earlier works~\cite{huang2004attacking,park2012identigram,pei2006novel}, the user's interaction is always required to remove watermark. Namely, they require the location of the watermark and recover that area. However, it can be not practical when processing massive images without location information. In~\cite{dekel2017effectiveness} and~\cite{gandelsman2019double}, they assume that the same watermark is added to all host images. Nevertheless, this assumption is also too strong to apply in real scenarios.

With the development of deep learning, neural networks show a great power in computer vision tasks~\cite{he2016deep,krizhevsky2012imagenet,zhao2017pyramid,goodfellow2014generative}. Several works try to apply neural networks to formulate an end-to-end problem, and there are two popular ways applied to solve it. One way is to directly formulate the watermark removal as an image-to-image translation task~\cite{cao2019generative,li2019towards}; the other way is adopting a two-stage strategy to formulate the problem: the first step is to locate the by a mask, and the second step is to recover the background in the watermark area and train a network to solve both at the same time~\cite{cun2021split,hertz2019blind,liu2021wdnet,liang2021visible}. The latter method was found in experiments can be more effective in watermark removal, so we mainly focus on preventing the second type of networks in this paper.

\vspace{2mm}
\noindent \textbf{Adversarial Attacks on Generative Models}. Szegedy \emph{et al.}~\cite{szegedy2013intriguing} first found and proposed the adversarial examples. In~\cite{goodfellow2014explaining}, Goodfellow \emph{et al.} proposed the Fast Gradient Sign Method (FGSM), which is a one-step gradient attack. After that, some stronger generation methods are proposed like I-FGSM~\cite{kurakin2016adversarial}, M-FGSM~\cite{dong2018boosting}, and Projected Gradient Descent (PGD)~\cite{madry2017towards}. In the black-box setting, adversarial examples can transfer across the different models~\cite{papernot2017practical,papernot2016transferability,dong2019evading,lin2019nesterov,xie2019improving} or can be generated by approximating gradients~\cite{chen2017zoo,uesato2018adversarial}. However, most of attack and defense works are about the classification problem\cite{Jia_2022_CVPR,jia2022boosting}.

Recent works have focused more on attacks on generative models. In~\cite{kos2018adversarial} and~\cite{tabacof2016adversarial}, the authors firstly explore adversarial attacks against  Variational Autoencoders (VAE) and VAE-GANS. Ruiz \emph{et al.}~\cite{ruiz2020disrupting,ruiz2020protecting} apply transferable adversarial attacks to disrupt facial manipulation systems. From then on some other works~\cite{yeh2020disrupting,chen2021magdr,segalis2020ogan,yang2021defending} adopt adversarial machine learning in translation-based deepfake models. These works show that adversarial attacks can defend against some malicious generative networks to protect users' privacy. Khachaturov \emph{et al.}~\cite{khachaturov2021markpainting} find that adversarial perturbations can fool an inpainting system into generating a patch similar to random pictures. Although this work can be applied to protect watermarks, the inpainting-based watermarking method needs obtaining a mask in image, and is not practical for big watermarks. Therefore, attacking the inpainting-based watermarking method~\cite{khachaturov2021markpainting} cannot protect the watermark in natural scenes.

\section{Methodology}

\subsection{Preliminary}
\label{sec:problem formulation}

The watermark vaccine generation can be formalized as an optimization problem with constraints. We assume a host image as \(x_h\), and the invisible watermark vaccine $\delta$ is injected onto it before adding watermark, which is restricted in $L_\infty$ norm bound $\epsilon$. Thus, we get the vaccinated image $\hat{x}_h$ by
\begin{equation}
\begin{array}{l}
\hat{x}_h = x_h+\delta \\
\|\delta\|_{\infty} \leq \varepsilon.
\end{array}
\end{equation}
Then we select a watermark sample $w$ from the watermark set $W$, where $w \in W$. The adding watermark operation can be assumed as $g$. And $g$ requires some parameters $\theta$ to identify the location $(p,q)$, the size $(u,v)$ and the transparency $\alpha$ of the watermark $w$ injected on the host image $x_h$. So a watermarked image $\hat{x}_{w}$ with vaccine can be defined as follows,
 
\begin{equation}
\begin{array}{l}
\hat{x}_{w}=g\left(\hat{x}_h, \omega, \theta\right),
\end{array}
\end{equation}
where $\theta=(p, q, u, v, \alpha)$. In practice, the parameters $p, q, n, m$ and $\alpha$ are always random. Similarly, the watermarked image without vaccine $x_{w}$ can also be obtained in the same way. Then we assume the blind watermark-removal network as $f$, and we can get the watermark removed images and masks of $x_{w}$ and $\hat{x}_{w}$ respectively through the network, which is defined as 
 \begin{equation}
\begin{array}{l}
X_w, M_w=f\left(x_{w}\right), \\
\hat{X}_w, \hat{M}_w=f\left(\hat{x}_{w}\right).
\end{array}
\end{equation}

Here, we denote the measurement of watermark removal effect as $Q(\cdot)$, and the goal of the vaccine is to degrade the removal effect of watermark removed images by minimizing $Q( f ( g(x_h + \delta, w, \theta) ))$. In addition, we desire our watermark vaccine to be universal for different watermark patterns, positions, sizes, transparencies, thus the expected $Q( f ( g(x_h + \delta, w, \theta) ))$ over different watermark $w$ and adding parameters $\theta$ is required to make vaccine watermark-agnostic. Our watermark vaccine generation can be formulated as follows,
\begin{equation}
\min_\delta \quad \mathbb{E}_{w \sim W}  \,   \mathbb{E}_{\theta \sim \Theta}\left[ Q( f ( g(x_h + \delta, w, \theta) )) \right].
\label{original_formulation}
\end{equation}

Unfortunately, there are two challenges in solving the above optimization problem. First, the effect of watermark removal $Q(\cdot)$ can be customized in a variety of different ways, but it is required to be differentiable during optimization. In addition, the two expectation over $W$ and $\Theta$ is hard to optimize by considering the loss of all combinations simultaneously. Although we can refer to `universal adversarial perturbation' in \cite{moosavi2017universal,hendrik2017universal,shafahi2020universal,mopuri2018ask}, it is still time-consuming and difficult to obtain the optimal vaccine. To address these issues, we further propose two types of watermark vaccine in the following.

\subsection{Disrupting Watermark Vaccine (DWV)}
\label{sec:dwv}
One way to protect the watermark is to disrupt the watermark-removed images, which means that as long as the watermarked image with the watermark vaccine passes through the watermark-removal networks, the output image will be ruined and could never be used. Thus, we call this vaccine as \textbf{Disrupting Watermark Vaccine (DWV)}. Next, in order to avoid the two expectations in Equation~\ref{original_formulation}, we decide to generate the watermark-agnostic vaccine on the host images instead of watermarked images. Therefore, we inject vaccine $\delta$ on the clean host image $x_h$ and get the vaccinated image $\hat{x}_h$. After passing through the network $f$, we get the watermark removed image and watermark removed mask, which is denoted as $\hat{X}_h$ and $\hat{M}_h$, because they are generated on clean host images without a watermark. Such operation can be formulated as,
\begin{equation}
\hat{X}_{h}, \hat{M}_h=f(\hat{x}_h),
\end{equation}
Then, we define the $\mathcal{L_{DWV}}$ to measure the distance between the watermark removed image $\hat{X}_{h}$ and the clean host image $x_h$,
\begin{equation}
\mathcal{L_{DWV}}\left( x_h,\delta \right) =\left\|\hat{X}_{h}-x_h\right\|^{2},
\label{dis}
\end{equation}
where the image distance adopt the mean-square error to measure.

The objective here is to maximize $\mathcal{L_{DWV}}$ such that the watermark removed image is significantly different from the host image. Thus, the watermark removed image of the host image is severely ruined by watermark-removal networks. Naturally, whatever watermark is added onto the image, the benign areas (the areas without the watermark) will be destroyed as well. As a result, the watermark removed images of watermarked images sufficiently deteriorate such that it has to be discarded or such that the modification is perceptually evident
The problem can be formally expressed as follows,
\begin{equation}
\begin{array}{cc}
\max _{\delta} & \mathcal{L_{DWV}}\left( x_h,\delta \right) \\
\text { s.t. }& \quad \hat{x}_h=x_h+\delta \\
&\|\delta\|_{\infty} \leqslant \varepsilon,\label{problem1}
\end{array}
\end{equation}

To be consistent with the previous requirements, we restrict the perturbations $\delta$ in $L_\infty$ norm bound $\epsilon$. We use projected gradient descent (PGD)~\cite{madry2017towards} to solve the optimization problem in Equation~\ref{eq:8}, and we can get the optimal $\delta$ according to the following iterative formula,
\begin{equation}
\delta^{t+1}=\operatorname{Proj}\left(\delta^{t}+\alpha \operatorname{sign}\left(\nabla_{\delta^{t}} \, \mathcal{L_{DWV}}\left( x_h,\delta^t \right)\right)\right),
\label{eq:8}
\end{equation}
where $\nabla_{\delta^{t}} \, \mathcal{L_{DWV}}\left( x_h,\delta^t \right)$ is the gradient of the disrupting loss w.r.t  $\delta^{t}$. $\alpha$ is the step size, $\operatorname{Proj()}$ denotes project the $\delta^{t}$ within the norm bound $(-\epsilon,\epsilon)$ and project the $x+\delta^{t}$ within the valid space $(0,1)$. In Fig.~\ref{framework} (a), we can see the framework of DWV generation, and at the bottom shows the inference of DWV. 

\subsection{Inerasable Watermark Vaccine (IWV)}
\label{sec:hwv}
Contrasted with DWV to protect the watermark by ruining the watermark removed images, another solution is to prevent the watermark from being identified and removed. To this end, as an alternative to DWV, we propose another vaccine in this section, \textbf{Inerasable Watermark Vaccine (IWV)}. It aims to make the watermarks hard to be detected and removed. As a result, the watermark patterns can not be erased completely on the watermark removed images. Inspired by the Equation~\eqref{dis}, we design the $\mathcal{L_{IWV}}$ as follow:
\begin{equation}
\mathcal{L_{IWV}}\left( x_h,\delta \right) =\frac{1}{2}\left(\beta\left\|\hat{X}_{h}-x_h\right\|^{2}+\|\hat{M}_h-\textbf{0}\|^{2}\right),
\label{hid}
\end{equation}
where $\hat{X}_{h}$ and $\hat{M}_h$ is the output of the blind watermark-removal network $f$, $x_h$ is the host image, \textbf{0} is a zero matrix, which is the same size as the predicted mask. There are two distance terms in the loss $\mathcal{L_{IWV}}$: image term and mask term. The image term is equal to the Equation~\eqref{dis}, and the mask term measures the distance between the predicted mask $\hat{M}_h$ and a zero matrix \textbf{0}. The $\beta$ is the hyperparameter to balance two loss terms.

Ideally, the predicted image $\hat{X}_{h}$ should be almost the same as the input image $x_h$, and the predicted mask $\hat{M}_h$ should be almost black, which means there is no watermark can be detected on $\hat{x}_h$ and the $\mathcal{L_{IWV}}$ should be very close to 0. However, in reality, these are not 0 and show a large loss actually, as shown in Fig.~\ref{framework}(b). Based on this situation, we decide to minimize the $\mathcal{L_{IWV}}$ to generate the vaccine that can make the output of the watermark-removal network close to the ideal one. This seems to be a well-intentioned fix for performance, but it is actually superfluous and adversarial. In the test stage, whatever watermark is added to the host image, the IWV will suppress the removal network to recognize it, and the outputs still preserve the watermarks and they tend to be the same as watermarked image inputs. Thus, the IWV generation can be formulated as, 

\begin{equation}
\begin{array}{cc}
\min _{\delta} & \mathcal{L_{IWV}}\left( x_h,\delta \right) \\
\text { s.t. }& \quad \hat{x}_h=x_h+\delta \\
&\|\delta\|_{\infty} \leqslant \varepsilon.
\label{problem2}
\end{array}
\end{equation}
We also restrict the perturbations $\delta$ in $L_\infty$ norm bound $\epsilon$, and solve it by projected gradient descent (PGD)~\cite{madry2017towards} again as follows,
\begin{equation}
\delta^{t+1}=\operatorname{Proj}\left(\delta^{t}-\alpha \operatorname{sign}\left(\nabla_{\delta^{t}} \,  \mathcal{L_{IWV}}\left( x_h,\delta \right)\right)\right).
\end{equation}

In Fig.~\ref{framework} (b), we can see the framework of IWV generation and the inference of IWV. The pseudocode of our algorithm to generate watermark vaccine including DWV and IWV is shown in Algorithm~\ref{algorithm}. We use projected gradient descent (PGD)~\cite{madry2017towards} to solve the problem~\eqref{problem1} or~\eqref{problem2}, and then we generate the DWV/IWV. During the inference stage, we inject the DWV and IWV onto host image and add the watermark on it. After that, we evaluate their adversarial results through the watermark-removal networks $f$ and expect to get the damaged or watermark-preserved image. In addition, we theoretically analyze why our vaccines work effectively and give the lower or upper bound of the watermark protection for DWV and IWV in Sec.1 of the Supplementary Material.

\begin{algorithm}[tb]
    \LinesNumbered
    \KwIn {host image $x_h$, blind watermark-removal network \(f\), iteration T, step size $\alpha$, perturbation bound $\epsilon$}
    \KwOut {Host image with watermark vaccine $\hat{x}_h$ }
    $\delta \leftarrow  0 , \hat{x}_h \leftarrow x_h+ \delta$ \;
    \For{i = 1 to T}{  
          \eIf{ vaccine is `DWV'}{
    using Equation \eqref{dis} to calculate the $\mathcal{L_{DWV}}$\;
    \(\delta\leftarrow \delta + \alpha \) sign$\left(\nabla_{\delta} \mathcal{L_{DWV}}\left( x_h,\delta \right) \right)$\;
      }{
      using Equation \eqref{hid} to calculate the $\mathcal{L_{IWV}}$\;
         \(\delta\leftarrow \delta - \alpha \) sign$\left(\nabla_{\delta}  \mathcal{L_{IWV}}\left( x_h,\delta \right)\right)$\;
      }
		$\hat{x}_h\leftarrow x_h + \text{clip}(\delta,-\epsilon,\epsilon)$\;
    }
    $\hat{x}_h\leftarrow \text{clip}(\hat{x}_h,0, 1)$\;
\caption{Watermark Vaccine Generation}
\label{algorithm}
\end{algorithm}

\section{Experiments}

\subsection{Experimental Setups}

\noindent \textbf{Datasets.}  We use the CLWD (Colored Large-scale Watermark Dataset)~\cite{liu2021wdnet} in our experiments, which contains three parts: watermark-free images, watermarks and watermarked images. We first pretrain the watermark-removal networks using watermarked images in the train set of CLWD. Then in the attack stage, we use the watermark-free images as host images to generate watermark vaccines, and then add the watermarks with generated watermark vaccines. The details about the dataset can be checked in Sec.2 of the Supplementary Material.

\noindent \textbf{Models Architectures.}  We choose three advanced blind watermark-removal networks: BVMR~\cite{hertz2019blind}, SplitNet~\cite{cun2021split}, and WDNet~\cite{liu2021wdnet}. We train them on the watermarked images of CLWD and save the best checkpoint parameters. 
 
\noindent \textbf{Evaluation Metrics.} 
Following the previous  work~\cite{liu2021wdnet,cun2021split,liang2021visible}, Peak Signal-to-Noise Ratio (PSNR), Structural Similarity (SSIM), Root-Mean-Square distance (RMSE), and weighted Root-Mean-Square distance (RMSE$_w$) are adopted as our evaluation metrics. The difference between RMSE and RMSE$_w$ is that RMSE$_w$ only focuses on the watermarked area. For DWV, we specify these metrics as PSNR$^h$, SSIM$^h$, RMSE$^h$, RMSE$^h_w$ compared with host images for a better illustration. The lower PSNR$^h$/SSIM$^h$ or the higher RMSE$^h$/RMSE$^h_w$ mean the worse results of watermark-removal networks thus the better protection performance of proposed DWV. 
For IWV, we specify the metrics as PSNR$^w$, SSIM$^w$, RMSE$^w$, RMSE$^w_w$ which are compared with watermarked images. Different from DWV, the higher PSNR$^w$/SSIM$^w$ or the lower RMSE$^w$/RMSE$^w_w$ mean that the more excellent performance on keeping the watermarks on thus the better protection performance of proposed IWV.
 
\noindent \textbf{Attack Parameters.} During attack, we empirically set the $L_\infty$ norm bound $\epsilon$ as 8/255, which is imperceptible by human eyes. We set the step size $\alpha$ as 2/255, and iteration $T$ as 50. We set the hyperparameter $\beta$ in Eq.~\ref{hid} for IWV to 2 initially. We also further discuss the sensitivity of these hyperparameters in the Sec.5 of Supplemental Material.
 
\subsection{Effectiveness of Watermark Vaccine}
\label{sec:exp result}

\begin{figure*}[tb]
\centering
\includegraphics[width=1\textwidth]{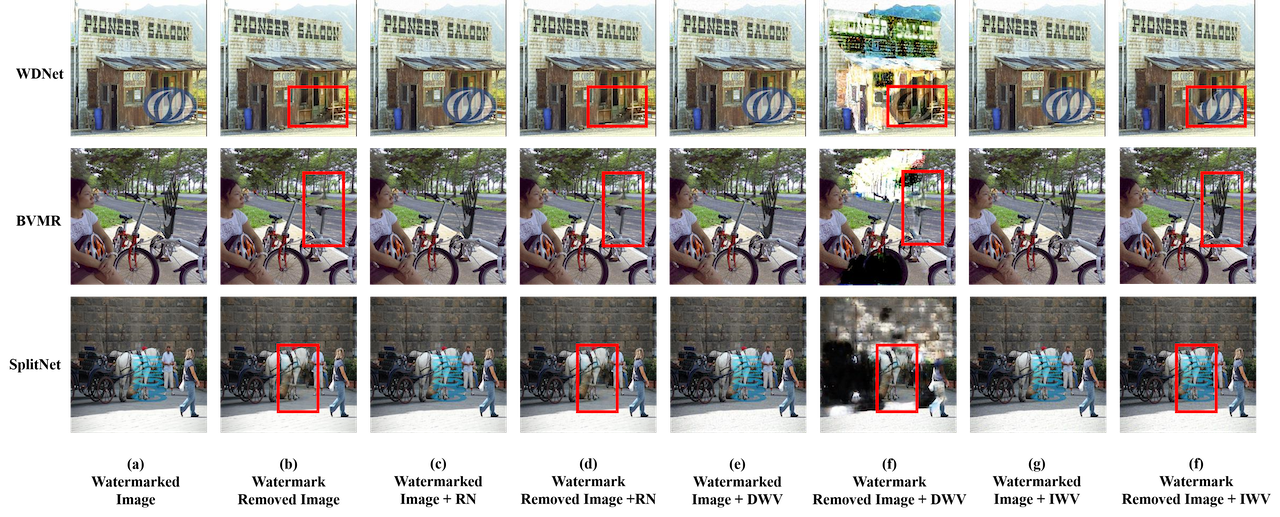}
\caption{Qualitative comparison of DWV and IWV. In each row, we show the watermark removal effects on the images without any vaccine, the images with random noises, the images with DWV, and the images with IWV under the same network.}
\label{visualize}
\end{figure*}

\begin{table}[tb]
\caption{Impact of the two vaccines on WDNet, BVMR, and SplitNet on the same dataset and with the same parameters $\theta$. The perturbations and random noises are restricted in $L_\infty$ norm bound 8/255. \textbf{Clean} denotes the watermarked image with no vaccines, and \textbf{RN} denotes the watermarked images with random noise. For DWV, the lower PSNR$^h$/SSIM$^h$ or the higher RMSE$^h$/RMSE$^h_w$ the better. For IWV, the higher PSNR$^w$/SSIM$^w$ or the lower RMSE$^w$/RMSE$^w_w$ the better. The best-protection results are denoted in boldface.}
\begin{minipage}{\linewidth}
\centering
\resizebox{\textwidth}{!}{
\begin{tabular}{@{}c|cccc|cccc|cccc@{}}
\toprule
\multicolumn{1}{l|}{} & \multicolumn{4}{c|}{WDNet\cite{liu2021wdnet}} & \multicolumn{4}{c|}{BVMR\cite{hertz2019blind}} & \multicolumn{4}{c}{SplitNet\cite{cun2021split}} \\ \midrule
Metrics & PSNR$^h$ & SSIM$^h$ & RMSE$^h$ & RMSE$^h_w$ & PSNR$^h$ & SSIM$^h$ & RMSE$^h$ & RMSE$^h_w$ & PSNR$^h$ & SSIM$^h$ & RMSE$^h$ & RMSE$^h_w$ \\ \midrule
Clean & 38.62 & 0.9946 & 3.09 & 16.25 & 41.96 & 0.9955 & 2.09 & 23.86 & 42.32 & 0.9939 & 2.12 & 21.86 \\
RN & 38.19 & 0.9938 & 3.23 & 17.06 & 42.48 & 0.9957 & 1.98 & 24.13 & 42.73 & 0.9943 & 2.07 & 21.33 \\
DWV(Ours) & \textbf{29.68} & \textbf{0.6360} & \textbf{8.47} & \textbf{41.36} & \textbf{29.43} & \textbf{0.6462} & \textbf{8.68} & \textbf{26.85} & \textbf{34.12} & \textbf{0.8951} & \textbf{5.18} & \textbf{67.68} \\ \bottomrule
\end{tabular}}
\centerline{}
\centerline{(a) The effect of DWV on different watermark-removal networks.}
\end{minipage}

\begin{minipage}{\linewidth}
\centering
\resizebox{\textwidth}{!}{
\begin{tabular}{@{}c|cccc|cccc|cccc@{}}
\toprule
\multicolumn{1}{l|}{} & \multicolumn{4}{c|}{WDNet\cite{liu2021wdnet}} & \multicolumn{4}{c|}{BVMR\cite{hertz2019blind}} & \multicolumn{4}{c}{SplitNet\cite{cun2021split}} \\ \midrule
Metrics & PSNR$^w$ & SSIM$^w$ & RMSE$^w$ & RMSE$^w_w$ & PSNR$^w$ & SSIM$^w$ & RMSE$^w$ & RMSE$^w_w$ & PSNR$^w$ & SSIM$^w$ & RMSE$^w$ & RMSE$^w_w$ \\ \midrule
Clean & 37.76 & 0.9788 & 3.42 & 52.77 & 41.88 & 0.9893 & 2.13 & 42.68 & 40.91 & 0.9788 & 2.53 & 49.67 \\
RN & 37.53 & 0.9755 & 3.50 & 52.95 & 42.59 & 0.9917 & 2.00 & 42.73 & 41.59 & 0.9795 & 2.41 & 49.29 \\
IWV(Ours) & \textbf{45.16} & \textbf{0.9831} & \textbf{2.24} & \textbf{28.00} & \textbf{43.31} & \textbf{0.9926} & \textbf{1.86} & \textbf{37.42} & \textbf{42.79} & \textbf{0.9834} & \textbf{2.23} & \textbf{35.00} \\ \bottomrule
\end{tabular}}
\centerline{}
\centerline{(b) The effect of IWV on different watermark-removal networks.}
\end{minipage}
\label{exp_results}
\end{table}

We evaluate the effectiveness of DWV and IWV on 10,000 random host-watermark image combinations. We test three models with the same watermark parameters $\theta$ on the same dataset. We compare the clean input and the input with random noise at the same time, which is also restricted in $L_\infty$ norm bound $(-\epsilon,\epsilon)$.

In Fig.~\ref{visualize}, we show the qualitative visualization of different networks and their corresponding results. It shows that no matter which network it is, the watermark removed images can be ruined if the host images are injected with DWV, although the watermarks can be successfully removed. For IWV, there are still noticeable all or part of watermarks on the watermark removed images, and other parts of the images are not damaged. On the contrary, the inputs with random noise present no protective effect on the watermark removed results for any watermark-removal network. We show more visualization results in Sec.3 of the Supplementary Material.

Tab.~\ref{exp_results} demonstrates the quantitative results of watermark vaccines on different watermark-removal networks. In Tab.~\ref{exp_results}(a), we can find that random noise could not disrupt the watermark removed image, while the DWV can significantly degrade the quality of watermarked removed images with the lower PSNR$^h$/SSIM$^h$ and the higher RMSE$^h$/RMSE$^h_w$ than others. On the other hand, by observing Tab.~\ref{exp_results}(b), the watermark removed images with IWV have a better similarity with watermarked input with a little higher PSNR/SSIM. It is noticeable that the RMSE$^w_w$ for IWV is much lower than others, which is to evaluate whether the watermark part is well preserved. Moreover, the above phenomena in different watermark-removal networks tend to be the same, and the quantitative results are consistent with the qualitative visualization. 

Although DWV can ruin the watermark removed images, the watermark patterns can also be removed. On the contrary, the watermarks with IWV could be still noticeable on the watermark removed images by human eyes. Therefore, which type of vaccine to choose depends on the need for protection. 

\subsection{Universality of Watermark Vaccine}

\begin{table}[ht]
\caption{Mean and standard deviation over evaluation metrics of DWV and IWV for random watermark patterns and location parameters.}
\begin{minipage}[t]{0.49\linewidth}
\resizebox{\linewidth}{!}{
\centering
\begin{tabular}{@{}c|cc|cc@{}}
\toprule
\multicolumn{1}{l|}{} & \multicolumn{2}{c|}{Watermark} & \multicolumn{2}{c}{Location} \\ \midrule
Metrics & \multicolumn{1}{c|}{Clean} & DWV & \multicolumn{1}{c|}{Clean} & DWV \\ \midrule
PSNR$^h$ & \multicolumn{1}{c|}{39.12$\pm$0.02} & 29.40$\pm$0.03 & \multicolumn{1}{c|}{40.82$\pm$0.04} & 28.95$\pm$0.01 \\
SSIM$^h$ & \multicolumn{1}{c|}{0.9957$\pm$0.0000} & 0.6021$\pm$0.0028 & \multicolumn{1}{c|}{0.9974$\pm$0.0001} & 0.5288$\pm$0.0020 \\
RMSE$^h$ & \multicolumn{1}{c|}{2.85$\pm$0.01} & 8.74$\pm$0.02 & \multicolumn{1}{c|}{2.37$\pm$0.01} & 9.15$\pm$0.01 \\
RMSE$^h_w$ & \multicolumn{1}{c|}{17.15$\pm$0.18} & 42.88$\pm$0.77 & \multicolumn{1}{c|}{16.67$\pm$0.11} & 52.02$\pm$0.68 \\ \bottomrule
\end{tabular}}
\centerline{}
\centerline{(a) DWV}
\end{minipage}
\hfill
\begin{minipage}[t]{0.49\linewidth}
\resizebox{\linewidth}{!}{
\centering
\begin{tabular}{@{}c|cc|cc@{}}
\toprule
\multicolumn{1}{l|}{} & \multicolumn{2}{c|}{Watermark} & \multicolumn{2}{c}{Location} \\ \midrule
Metrics & \multicolumn{1}{c|}{Clean} & IWV & \multicolumn{1}{c|}{Clean} & IWV \\ \midrule
PSNR$^w$ & \multicolumn{1}{c|}{38.42$\pm$0.02} & 47.35$\pm$0.21 & \multicolumn{1}{c|}{40.37$\pm$0.03} & 52.30$\pm$0.30 \\
SSIM$^w$ & \multicolumn{1}{c|}{0.9874$\pm$0.002} & 0.9938$\pm$0.0003 & \multicolumn{1}{c|}{0.9956$\pm$0.0001} & 0.9981$\pm$0.0001 \\
RMSE$^w$ & \multicolumn{1}{c|}{3.08$\pm$0.01} & 1.63$\pm$0.03 & \multicolumn{1}{c|}{2.48$\pm$0.01} & 1.08$\pm$0.03 \\
RMSE$^w_w$ & \multicolumn{1}{c|}{54.25$\pm$0.54} & 20.67$\pm$0.39 & \multicolumn{1}{c|}{48.28$\pm$0.38} & 10.39$\pm$0.55 \\ \bottomrule
\end{tabular}}
\centerline{}
\centerline{(b) IWV.}
\end{minipage}
\label{table:univer1}
\end{table}

\begin{table}[ht]
\caption{The evaluation metrics for the Clean/DWV/IWV under different size and transparency of watermarks. Each row shows the results under different watermark sizes or different transparencies. The best-attacking results are denoted in boldface.}
\resizebox{\textwidth}{!}{
\begin{tabular}{@{}c|cc|cc|cc|cc||cc|cc|cc|cc@{}}
\toprule
Metrics & \multicolumn{2}{c|}{PSNR$^h$} & \multicolumn{2}{c|}{SSIM$^h$} & \multicolumn{2}{c|}{RMSE$^h$} & \multicolumn{2}{c||}{RMSE$^h_w$} & \multicolumn{2}{c|}{PSNR$^w$} & \multicolumn{2}{c|}{SSIM$^w$} & \multicolumn{2}{c|}{RMSE$^w$} & \multicolumn{2}{c}{RMSE$^w_w$} \\ \midrule
Input & Clean & DWV & Clean & DWV & Clean & DWV & Clean & DWV & Clean & IWV & Clean & IWV & Clean & IWV & Clean & IWV \\ \midrule
Size=60 & 39.91 & \textbf{29.16} & 0.9967 & \textbf{0.5610} & 2.62 & \textbf{8.95} & 18.02 & \textbf{48.87 }& 39.36 & \textbf{50.28} & 0.9927 & \textbf{0.9968} & 2.76 & \textbf{1.31} & 51.55 & \textbf{13.70} \\
Size=70 & 39.53 & 29.25 & 0.9962 & 0.5784 & 2.72 & 8.86 & 17.73 & 45.59 & 38.87 & 50.04 & 0.9901 & 0.9958 & 2.92 & 1.35 & 53.31 & 16.14 \\
Size=80 & 39.14 & 29.25 & 0.9957 & 0.5963 & 2.84 & 8.78 & 17.03 & 41.13 & 38.39 & 47.52 & 0.9868 & 0.9936 & 3.09 & 1.63 & 55.16 & 20.71 \\
Size=90 & 38.67 & 29.54 & 0.9950 & 0.6261 & 3.00 & 8.58 & 17.02 & 38.11 & 37.82 & 45.69 & 0.9833 & 0.9911 & 3.29 & 1.90 & 56.35 & 26.03 \\
Size=100 & 38.25 & 29.67 & 0.9945 & 0.6455 & 3.15 & 8.47 & 16.81 & 37.32 & 37.32 & 43.06 & 0.9792 & 0.9864 & 3.49 & 2.34 & 57.29 & 32.87 \\ \midrule
$\alpha$=0.45 & 39.24 & \textbf{29.31} & 0.9961 & \textbf{0.5984} & 2.81 & \textbf{8.80} & 15.51 & \textbf{49.44} & 38.35 & \textbf{48.28} & 0.9896 & \textbf{0.9948} & 3.10 & \textbf{1.52} & 45.92 & \textbf{18.00} \\
$\alpha$=0.50 & 39.19 & 29.43 & 0.9959 & 0.6129 & 2.83 & 8.70 & 16.28 & 44.32 & 38.36 & 46.27 & 0.9883 & 0.9940 & 3.10 & 1.73 & 50.69 & 20.67 \\
$\alpha$=0.55 & 39.14 & 29.25 & 0.9957 & 0.5963 & 2.84 & 8.78 & 17.03 & 41.13 & 38.39 & 47.52 & 0.9868 & 0.9936 & 3.09 & 1.63 & 55.16 & 20.71 \\
$\alpha$=0.60 & 39.08 & 29.37 & 0.9955 & 0.6004 & 2.86 & 8.75 & 17.79 & 39.26 & 38.34 & 47.70 & 0.9855 & 0.9934 & 3.10 & 1.59 & 59.44 & 22.14 \\
$\alpha$=0.65 & 38.99 & 29.32 & 0.9952 & 0.6048 & 2.89 & 8.79 & 18.53 & 39.31 & 38.27 & 47.19 & 0.9841 & 0.9934 & 3.13 & 1.54 & 62.54 & 21.40 \\ \bottomrule
\end{tabular}}
\label{table:univer2}
\end{table}

As mentioned in Sec.~\ref{sec:dwv} and~\ref{sec:hwv}, the watermark vaccine we proposed can adapt to different watermarks and parameters and has a good universality. To illustrate this, we investigate the different watermark patterns, sizes, locations and transparency of watermarks for 1,000 host images. The WDNet~\cite{liu2021wdnet} is selected as the model for an example. Other models can be found in the Supplementary Material. We test every host image with ten random-selected watermark patterns and fix other parameters. Similarly, we test every host image with ten random locations with a fixed watermark and other fixed parameters for the location. To show their universality, we calculate the mean and variance of these results from different settings. Concerning about the watermark size and transparency, we select six sizes: $60 \times 60 $, 70 $\times$ 70, 80 $\times$ 80, 90 $\times$ 90, 100 $\times$ 100 and six transparency parameters: $\alpha$ = 0.45, 0.50, 0.55, 0.60, 0.65. We  fix the $\alpha$ = 0.55 if the size varies, and fix size $= 80 \times 80$, if the transparency varies. Finally, we calculate their evaluation metrics respectively. The above quantitative results are shown in Tab.~\ref{table:univer1} and~\ref{table:univer2}.

In Tab.~\ref{table:univer1}, compared to the means of the clean input, the means of DWV and IWV show that our watermark vaccines are still effective, and the minor variances of vaccines prove that our vaccines can be universal among different watermark patterns or locations. Tab.~\ref{table:univer2} shows that the metrics in each row are not much different, although they are under different sizes and transparencies of the watermark. Hence, the above results indicate that DWV and IWV have good universality, regardless of the watermark pattern, position, size and transparency. The visualization of universality can be seen in the Sec.6 of Supplementary Material.

Interestingly, according to Tab. \ref{table:univer2}, we find that when the size of the watermark becomes larger, the performance of the DWV and IWV has dropped. Moreover, if the transparency parameter $\alpha$ of the watermark becomes larger, the effect of the protection will be worsen, especially for IWV. This phenomenon is consistent with our analysis in Sec.1 of Supplementary Material, that the watermarking variation $\|w\|$ is one of the factors that determine the effectiveness of watermark protection. The better performance of the watermark vaccine depends on a smaller variation of $\|w\|$. Therefore, it can be a challenge for the copyright owners to choose a suitable size and transparency for the watermark, where a larger and low-transparency watermark is convenient for copyright identification. In comparison, a smaller and high-transparency watermark is more beneficial to protect the watermark vaccine. 

\subsection{Transferability of Watermark Vaccine}

\begin{table}[tbp]
\caption{Vaccines Transferability. The columns correspond to the target model, while the rows correspond to the source model. For brevity, we show the RMSE$^h$ for DWV and the RMSE$^w_w$ for IWV.}
\begin{minipage}[t]{0.49\linewidth}
\centering
\resizebox{0.8\linewidth}{!}{
\begin{tabular}{@{}cc|ccc@{}}
\toprule
\multicolumn{2}{c|}{Target Model} & WDNet & BVMR & SplitNet \\ \midrule
\multicolumn{2}{c|}{Clean} & 3.14 & 2.78 & 2.25 \\
\multicolumn{2}{c|}{RN} & 3.29 & 2.70 & 2.22 \\ \midrule
\multicolumn{1}{c|}{\multirow{3}{*}{\begin{tabular}[c]{@{}c@{}}Source\\ Model\end{tabular}}} & WDNet & \textbf{8.30} & 2.80 & 2.34 \\
\multicolumn{1}{c|}{} & BVMR & 3.35 & \textbf{8.61} & 2.47 \\
\multicolumn{1}{c|}{} & SplitNet & 3.23 & 2.79 & \textbf{5.17} \\ \bottomrule
\end{tabular}}
\centerline{(a) RMSE$^h$ of DWV}
\end{minipage}
\hfill
\begin{minipage}[t]{0.49\linewidth}
\centering
\resizebox{0.8\linewidth}{!}{
\begin{tabular}{@{}cc|ccc@{}}
\toprule
\multicolumn{2}{c|}{Target Model} & WDNet & BVMR & SplitNet \\ \midrule
\multicolumn{2}{c|}{Clean} & 54.47 & 43.84 & 51.10 \\
\multicolumn{2}{c|}{RN} & 54.63 & 43.85 & 51.15 \\ \midrule
\multicolumn{1}{c|}{\multirow{3}{*}{\begin{tabular}[c]{@{}c@{}}Source\\ Model\end{tabular}}} & WDNet & \textbf{30.14} & 43.80 & 50.80 \\
\multicolumn{1}{c|}{} & BVMR & 54.38 & \textbf{40.13} & 50.93 \\
\multicolumn{1}{c|}{} & SplitNet & 54.52 & 43.50 & \textbf{37.87} \\ \bottomrule
\end{tabular}}
\centerline{(b) RMSE$^w_w$ of IWV}
\end{minipage}
\label{table:trans}
\end{table}

\begin{figure}[htbp]
\centering
\begin{minipage}{0.49\linewidth}
\centering
\includegraphics[width=0.8\textwidth]{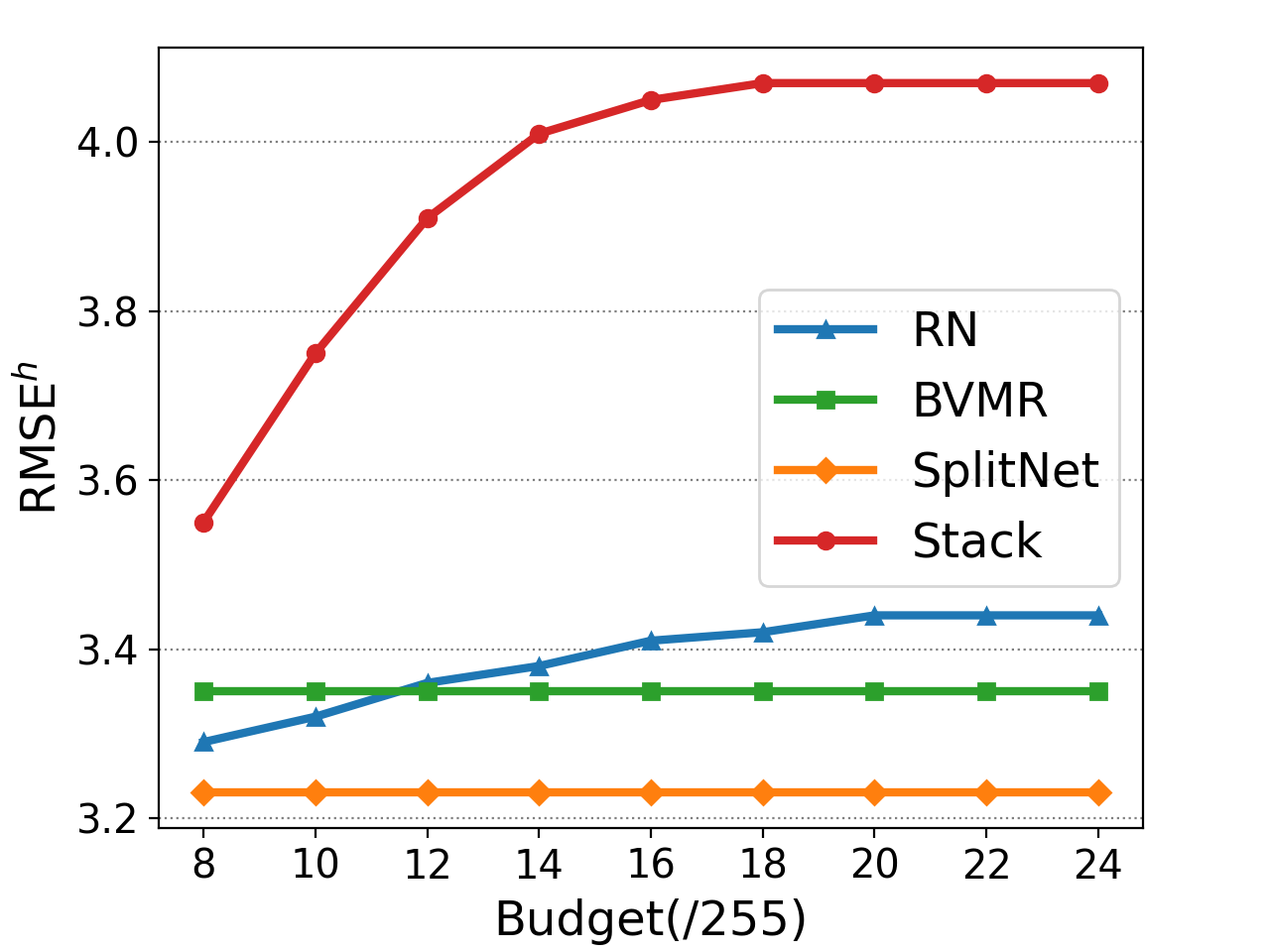}
\centerline{\footnotesize (a) RMSE$^h$ of DWV}
\end{minipage}
\begin{minipage}{0.49\linewidth}
\centering
\includegraphics[width=0.8\textwidth]{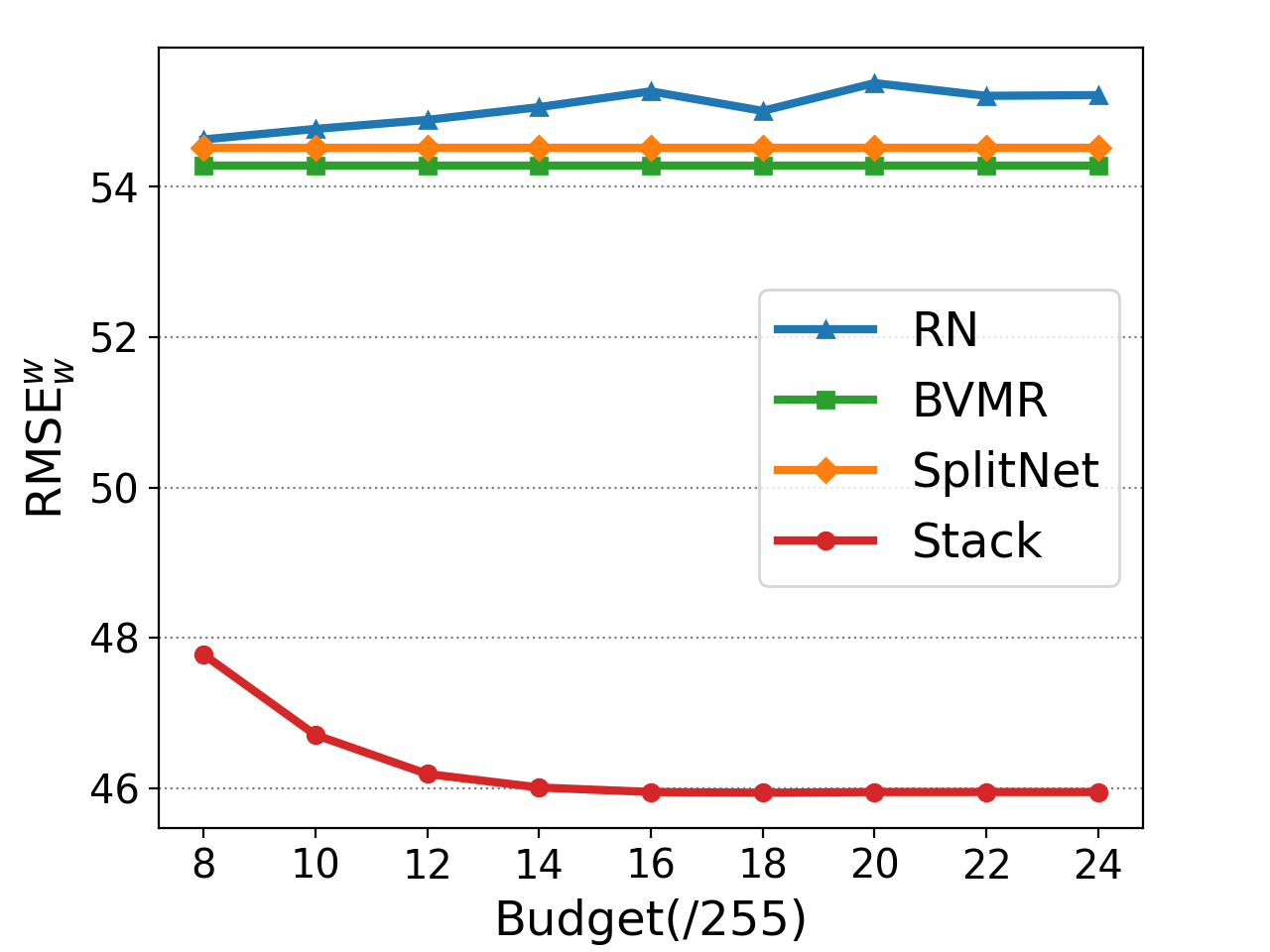}
\centerline{\footnotesize (b) RMSE$^w_w$ of IWV}
\end{minipage}
\caption{Testing stacked vaccines on WDNet~\cite{liu2021wdnet}. For a comparison, we first add a random perturbation baseline and also test the vaccines generated by BVMR and SplitNet respectively. The stacked ones perform clearly the best.}
\label{fig:stack}
\end{figure}

First, we explore the transferability of our vaccines across different watermark removal networks, and Tab.~\ref{table:trans} shows the results. We find that the vaccines show limited transferalibity across different watermark removal methods, which is the common problem of the adversarial examples on generative models. The reason may be related to the different procedures and network structures of removal networks. e.g., BVMR~\cite{hertz2019blind} is a one-stage method of predicting watermark removed images, while SplitNet~\cite{cun2021split} and WDNet~\cite{liu2021wdnet} contain detection, removal and refinement steps. In addition, SplitNet~\cite{cun2021split} adopts stacked attention-guided ResUNets, but other models do not. 

In real world, humans can be protected against different kinds of viruses by inoculating different vaccines. Inspired by this, we study the stacked vaccine assembled by three networks. We test the stacked vaccine on different watermark removal networks (see the partial test results on WDNet in Fig.~\ref{fig:stack}). Compared to the random perturbation baseline and the vaccines generated by other source models, the stacked vaccines perform better under various perturbation budgets. In future work, we will explore how to improve the transferability across different watermark removal networks/frameworks.

\subsection{Resistance to Image Processing Operations}

\begin{figure}[t]
\begin{minipage}{0.49\linewidth}
\centering
  \includegraphics[width=0.9\textwidth]{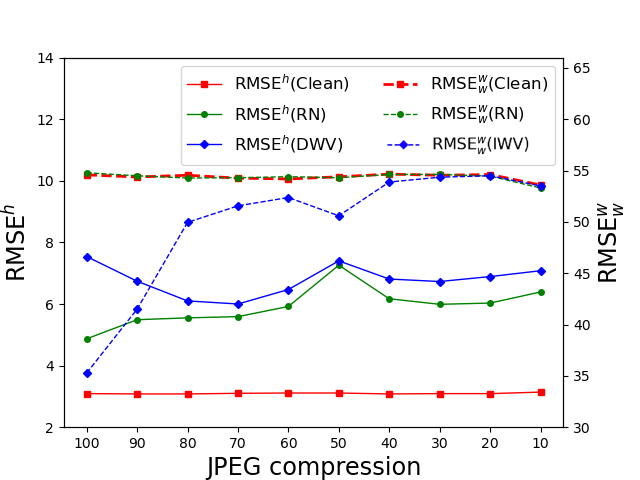}
\end{minipage}
\hfill
\begin{minipage}{0.49\linewidth}
\centering
  \includegraphics[width=0.9\textwidth]{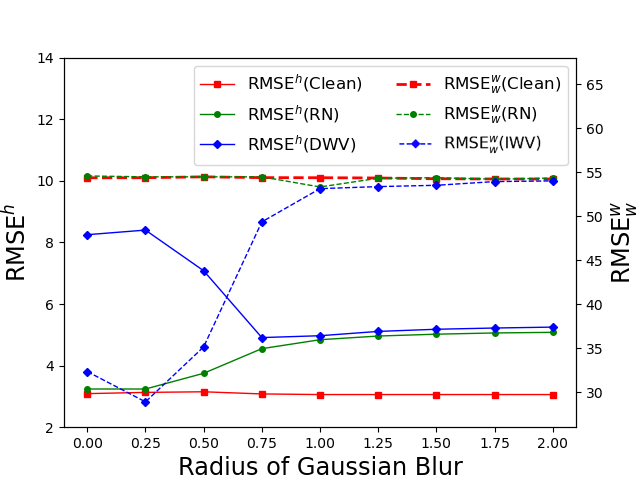}
\end{minipage}
\caption{Effect of two image-based transformation operations (JPEG Compression, Blur) on watermark vaccine. The solid lines show the change of RMSE$^h$, while the dashed lines show the RMSE$^w_w$ change. }
\label{fig:resist}
\end{figure}

In this section, we explore whether our vaccine can resist the common image processing operation. We select two common transformations: JPEG compression \cite{liu2019feature} and Gaussian blur \cite{cohen2019certified}, and take the WDNet as an example for the model. We average the results of 1,000 watermarked images and plot them as Fig. \ref{fig:resist}. For brevity, we only show the RMSE$^h$  of DWV and the RMSE$^w_w$ of IWV, and other metrics can be found in Sec.7 of the Supplementary Material. 

In Fig. \ref{fig:resist} (a), as the degree of JPEG compression ratio increases, it shows that the RMSE$^h$ for DWV is declined and higher than random noise at first. Then it gradually rises and approaches the variation of random noise finally. It is possibly because the performance degradation of the watermark vaccine is stronger than the image deterioration at the early stage, and when the degradation is strong enough, the result of DWV is similar to random noise. Regarding the IWV, we can find that the RMSE$^w_w$ of IWV has a sharp rise when the compression ratio increases, then it flattens out. It is worth noting that our watermark vaccines still have effects if the compression ratio is less than 80. The phenomenon in Gaussian blur is quite the same as that in JPEG compression in Fig. \ref{fig:resist}(b), and our watermark vaccines can resist the blur operation if the radius of Gaussian blur is less than 0.75. Besides the two image processing operations described above, we also consider some other operations that may affect our watermark vaccines, which will be present in the supplementary material. In conclusion, although some image-based transformation operations could reduce the effect of the watermark vaccine if their degradation is too substantial, they could also result in a lower quality of the image. Therefore, to some degree, our watermark vaccine can effectively resist some image processing operations.

\section{Conclusions}
Watermarking is an important and effective tool to protect copyright yet in face of the watermark removal threat. In this paper, we develop an idea of a watermark vaccine to protect watermarks. Our watermark vaccine is obtained by optimizing adversarial perturbations to attack the blind watermark removal network. 
Specifically, we propose two types of vaccine, dubbed disrupting watermark vaccine (DWV) and inerasable watermark vaccine (IWV). When malicious removal is presented, DWV will bring catastrophic damage to the host image, while IWV will keep the watermarks still clearly noticeable to human eyes. Both theoretical analysis and empirical experiments show that our vaccines is universal to different watermark patterns, sizes, locations, and transparencies, and they can also resist typical image transformation operations to a certain extent. This work makes the first exploration to protect watermarks from malicious removal. There is still space to improve our approach, e.g. by improving the transferability of watermark vaccines across target models. We leave further explorations in future work.

~\\
\noindent \textbf{Acknowledgement}\\
Supported by the National Key R$\&$D Program of China under  (Grant 2019YFB \\
1406500), Sponsored by Ant Group Security and Risk Management Fund.

\clearpage
%
%
\bibliographystyle{splncs04}
\bibliography{egbib}
\end{document}